\documentclass[10pt,twocolumn,letterpaper]{article}

\usepackage{cvpr}
\usepackage{times}
\usepackage{epsfig}
\usepackage{graphicx}
\usepackage{amsmath}
\usepackage{amssymb}

\usepackage{algorithm}
\usepackage{algorithmic}
\usepackage{varwidth}
\usepackage{makecell}
\usepackage{graphicx}
\usepackage{subfig}
\usepackage{comment}
\newcolumntype{K}[1]{>{\centering\arraybackslash}p{#1}}
\let\emptyset\varnothing

\usepackage[breaklinks=true,bookmarks=false]{hyperref}

\cvprfinalcopy 

\def\cvprPaperID{1019} 

\usepackage[usenames,dvipsnames,svgnames,table]{xcolor}
\newcommand{\mohamed}[1]{{\color{blue}{( #1)}}}

\newcommand{\ethan}[1]{{\color{red}{( #1)}}}

\begin{document}

\title{A Generative Adversarial  Approach for Zero-Shot Learning from Noisy Texts}
\author{Yizhe Zhu$^{1}$, \quad Mohamed Elhoseiny$^{2}$, \quad  Bingchen Liu$^{1}$, \quad Xi Peng$^{1}$ \quad and Ahmed Elgammal$^{1}$ \\
	  yizhe.zhu@rutgers.edu,  \quad elhoseiny@fb.com, \\ \quad \{bingchen.liu, xipeng.cs\}@rutgers.edu, \quad elgammal@cs.rutgers.edu 
	  \\
	$^{1}$Rutgers University, Department of Computer Science,  $^{2}$ Facebook AI Research 
}


\maketitle

\begin{abstract}


Most existing zero-shot learning methods consider the problem as a visual semantic embedding one. Given the demonstrated capability of Generative Adversarial Networks(GANs) to generate images, we instead leverage GANs to imagine unseen categories from text descriptions and hence recognize novel classes with no examples being seen. Specifically, we propose a simple yet effective generative model that takes as input noisy text descriptions about an unseen class (e.g.Wikipedia articles) and generates synthesized visual features for this class. With added pseudo data, zero-shot learning is naturally converted to a traditional classification problem. Additionally, to preserve the inter-class discrimination of the generated features, a visual pivot regularization is proposed as an explicit supervision.
Unlike previous methods using complex engineered regularizers, our approach can suppress the noise well without additional regularization. Empirically, we show that our method consistently outperforms the state of the art on the largest available benchmarks on Text-based Zero-shot Learning. 

 
\end{abstract}

\section{Introduction}
In the conventional object classification tasks, samples of all classes are available for training a model. However, objects in the real world have a long-tailed distribution.
In spite that images of common concepts can be readily found, there remains a tremendous number of concepts with insufficient and sparse visual data, thus making the conventional object classification methods infeasible.
Targeting on tackling such an unseen object recognition problem, zero-shot learning has been widely researched recently.

\begin{figure}[t!]
	\centering
	\includegraphics[width=8.5cm]{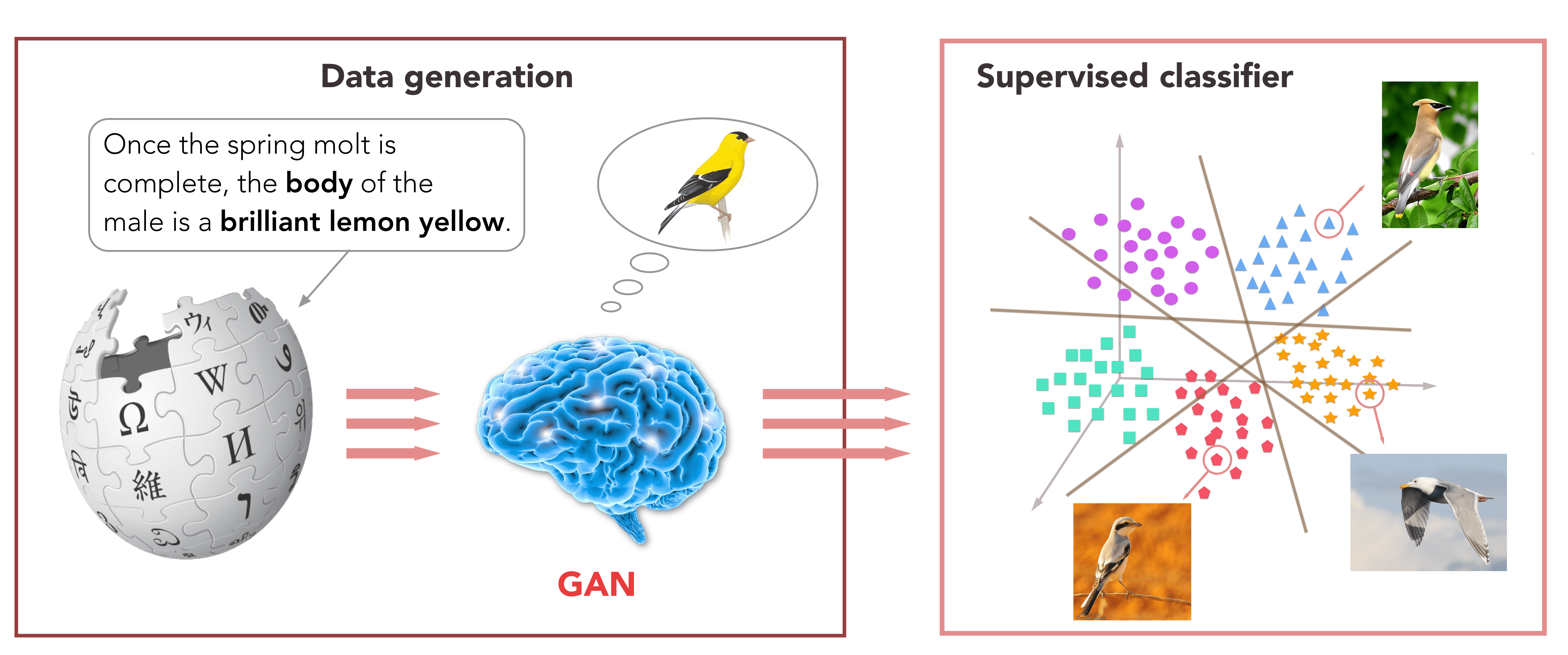}
	\caption{Illustration of our proposed approach. We leverages GANs to visually imagine the objects given noisy Wikipedia articles. With hallucinated features, a supervised classifier is trained to predict image's label.}
	\label{fig:intro}
	\vspace{-6.5mm}
\end{figure}

The underlying secret ensuring the success of zero-shot learning is to find an intermediate semantic representation (e.g. attributes or textual features) to transfer the knowledge learned from seen classes to unseen ones~\cite{Elhoseiny_2017_CVPR}. The majority of state-of-the-art approaches~\cite{akata2016label, akata2015evaluation, romera2015embarrassingly, zhang2016learning, socher2013zero, frome2013devise, tsai2017learning} consider zero-shot learning as a visual-semantic embedding problem. The paradigm can be generalized as training mapping functions that project visual features and/or semantic features to a common embedding space. The class label of an unseen instance is predicted by ranking the similarity scores between semantic features of all unseen classes and the visual feature of the instance in embedding space. 
Such a strategy conducts a one-to-one projection from semantic space to visual space. However, textual descriptions for categories and objects are inherently mapped to a variety of points in the image space. For example, ``a blue bird with white head'' can be the description of all birds with a blue body and a white head. This motivates us to study how adversarial training learns a one-to-many mapping with adding stochasticity. In this paper, we propose a generative adversarial approach for zero-shot learning that outperforms the state of the art by 6.5\% and 5.3\% on \emph{Caltech UCSD Birds-2011}(CUB)~\cite{wah2011caltech} and \emph{North America Birds}(NAB)~\cite{Horn2015} datasets respectively. 


\begin{figure*}
	\centering
    \includegraphics[width= 7.0in,height= 1.8in]{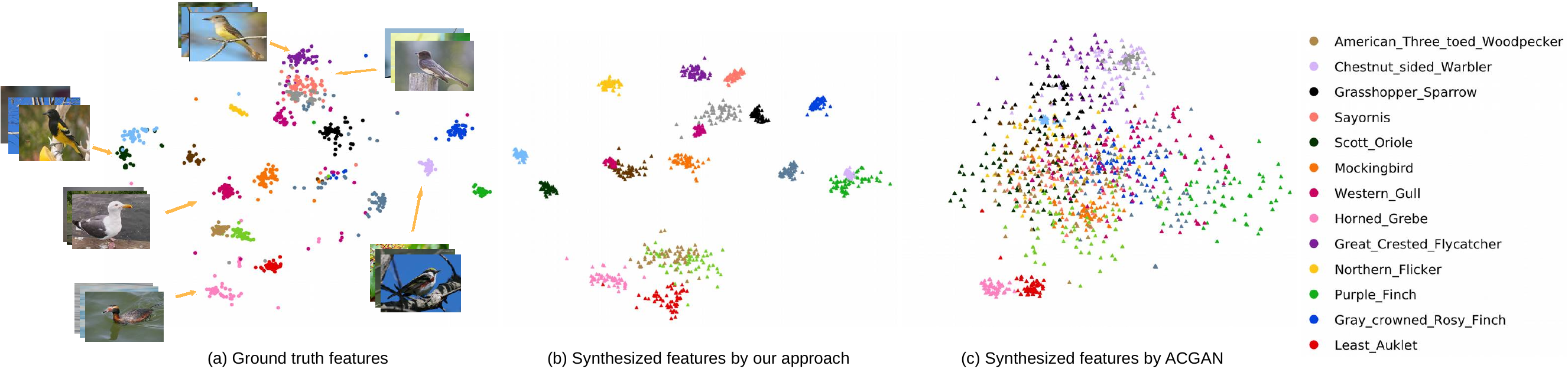}

	\setlength\belowcaptionskip{-3.0ex}
	\caption{t-SNE visualization of features from randomly selected unseen classes. The color indicates different class labels. Groundtruth features are marked as circles and synthesized ones as triangles. Our proposed method provides the intra-class diversity while preserving inter-class discrimination.}
	\label{fig:tSNE}
\end{figure*}

In this paper, we adopt a novel strategy that casts zero-shot learning as an imagination problem as shown in Fig.~\ref{fig:intro}. We focus on investigating how to hallucinate competent data instances that provide the intra-class diversity while keeping inter-class discrimination for unseen novel classes. Once this pseudo data is generated, a supervised classifier is directly trained to predict the labels of unseen images.

Recent years witness the success of generative adversarial networks (GANs)~\cite{goodfellow2014generative} to generate high compelling images. Our approach leverages GANs as a powerful computational model to imagine how unseen objects look like purely based on textual descriptions. Specifically, we first extract the semantic representation for each class from the Wikipedia articles. The proposed conditional generative model then takes as input the semantic representations of classes and hallucinates the pseudo visual features for corresponding classes. 
Unlike previous methods~\cite{guo2017synthesizing, long2017zero}, our approach does not need any prior assumption of feature distribution and can imagine an arbitrary amount of plausible features indefinitely. The idea is conceptually simple and intuitive, yet the proper design is critical. 

Unlike attributes consisting of the discriminative properties shared among categories, Wikipedia articles are rather noisy as most words are irrelevant to visually recognizing the objects. 
Realizing that noise suppression is critical in this scenario, previous methods~\cite{Elhoseiny_2017_CVPR,Qiao2016, romera2015embarrassingly, tsai2017learning} usually involve complex designs of regularizers, such as $L_{2,1}$ norm in ~\cite{Qiao2016, Elhoseiny_2017_CVPR} and autoencoder in ~\cite{tsai2017learning}.  In this work, we simply pass textual features through additional fully connected (FC) layer before feeding it into the generator.  
We argue that this simple modification achieves the comparable performance of noise suppression and increases the ZSL performance of our method by $\sim 3\%$($40.85\% $vs.$ 43.74\%$)  on CUB dataset. 
Besides, the sparsity of training data($\sim60$ samples per class in CUB) makes GANs alone hardly simulate well the conditional distribution of the high dimensional feature ($\sim 3500D$).
 As shown in Fig.~\ref{fig:tSNE}.c, the generated features disperse enormously and destroy the cluster structure in real features, thus hardly preserving enough discriminative information across classes to perform unseen image classification. To remedy this limitation, we proposed a visual pivot regularizer to provide an explicit guide for the generator to synthesize features in a proper range, thus preserving enough inter-class discrimination. Empirically, it aligns well the generated features as shown in Fig.~\ref{fig:tSNE}.b, and boosts the ZSL performance of our method from $22.83\%$ to $43.74\%$ on CUB.


Succinctly, our contributions are three-fold:

\textbf{1)} We propose a generative adversarial approach for ZSL (GAZSL) that convert ZSL to a conventional classification problem, by synthesizing the missing features for unseen classes purely based on the noisy Wikipedia articles.

\textbf{2)} 
We present two technical contributions: additional FC layer to suppress noise, and visual pivot regularizer to provide a complementary cue for GAN to simulate the visual distribution with greater inter-class discrimination. 


\textbf{3)} We apply the proposed GAZSL to multiple tasks, such as zero-shot recognition, generalized zero-shot learning, and zero-shot retrieval, and it consistently outperforms state-of-the-art methods on several benchmarks.

\section{Related Work}
\noindent\textbf{Zero-Shot Learning Strategy}
As one of the pioneering works, Lampert \emph{et al.}~\cite{lampert2009} proposed a Direct Attribute Prediction (DAP) model that assumed independence of attributes and estimated the posterior of the test class by combining the attribute prediction probabilities. 
Without the independence assumption, Akata \emph{et al.}~\cite{akata2016label} proposed an Attribute Label Embedding(ALE) approach that considers attributes as the semantic embedding of classes and thus tackles ZSL as a visual semantic embedding problem. Consequently, the majority of state-of-the-art methods converges to embedding-based methods.  The core of such approaches is to (a) learn a mapping function from the visual feature space to the semantic space~\cite{socher2013zero, frome2013devise, fu2016semi}, or conversely~\cite{zhang2016learning, shigeto2015ridge}, (b) or jointly learn the embedding function between the visual and semantic space through a latent space~\cite{yang2014unified, lei2015predicting, akata2015evaluation, romera2015embarrassingly, akata2016multi}. 

Apart from the aforementioned methods, a new strategy converted the zero-shot recognition to a conventional supervised classification problem by generating pseudo samples for unseen classes~\cite{guo2017synthesizing, long2017zero, guo2017zero}.
Guo \emph{et al.}~\cite{long2017zero} assumed a Gaussian distribution prior for visual features of each class and estimated the distribution of unseen class as a linear combination of those of seen classes.
Long \emph{et al.}~\cite{long2017zero} retained one-to-one mapping strategy and synthesized visual data via mapping attributes of classes or instances to the visual space. The number of synthesized data is rigidly restrained by the size of the dataset. Guo \emph{et al.}~\cite{guo2017zero} drew pseudo images directly from seen classes that inevitably introduces noise and bias. In contrast, our approach does not need any prior assumption of data distribution and can generate an arbitrary amount of pseudo data.

\noindent\textbf{Semantic representations}
Zero-shot learning tasks require leveraging side information as semantic representations of classes. Human specified attributes are popularly utilized as the semantic representation ~\cite{lampert2009, Lampert2014, akata2016label,frome2013devise, zhang2016learning, yang2014unified, lei2015predicting, akata2015evaluation, romera2015embarrassingly, akata2016multi}.  Despite the merit of attributes that provide a less-noisy and discriminative description of classes, the significant drawback is that attributes require being manually defined and collected, and field experts are often needed for such annotation, especially in fine-grained datasets~\cite{wah2011caltech, Horn2015}. 

Many researchers seek handier yet effective semantic representations based on class taxonomies~\cite{jiang1997semantic, resnik1995using} or text descriptions~\cite{elhoseiny2013write,Elhoseiny_2017_CVPR, Qiao2016, reed2016learning, lei2015predicting}. 
Compared with class taxonomies, text descriptions(e.g., Wikipedia articles) are more expressive and distinguishable. However, Wikipedia articles are rather noisy with superfluous information irrelevant to visual images. In this scenario, TF-IDF features~\cite{salton1988term} are commonly used for textual representation~\cite{elhoseiny2013write, Qiao2016, Elhoseiny_2017_CVPR} due to its superior performance.  Elhoseiny \emph{et al.}~\cite{elhoseiny2013write}
proposed an approach to that combines domain transfer and regression to predict visual classifiers from a TF-IDF textual representation. 
Qiao \emph{et al.}~\cite{Qiao2016} suppressed the noise in the text descriptions by encouraging group sparsity on the connections to the textual
terms. More recently, Elhoseiny \emph{et al.}~\cite{Elhoseiny_2017_CVPR} proposed a learning framework that is able to connect text terms to its relevant parts of objects and suppress connections to non-visual text terms without any part-text annotations.
Our method also leverages TF-IDF features while comparably suppressing the non-visual information without complicated regularizations.

\begin{figure*}[t!]
	\centering
	\includegraphics[width=15cm,  height=6cm]{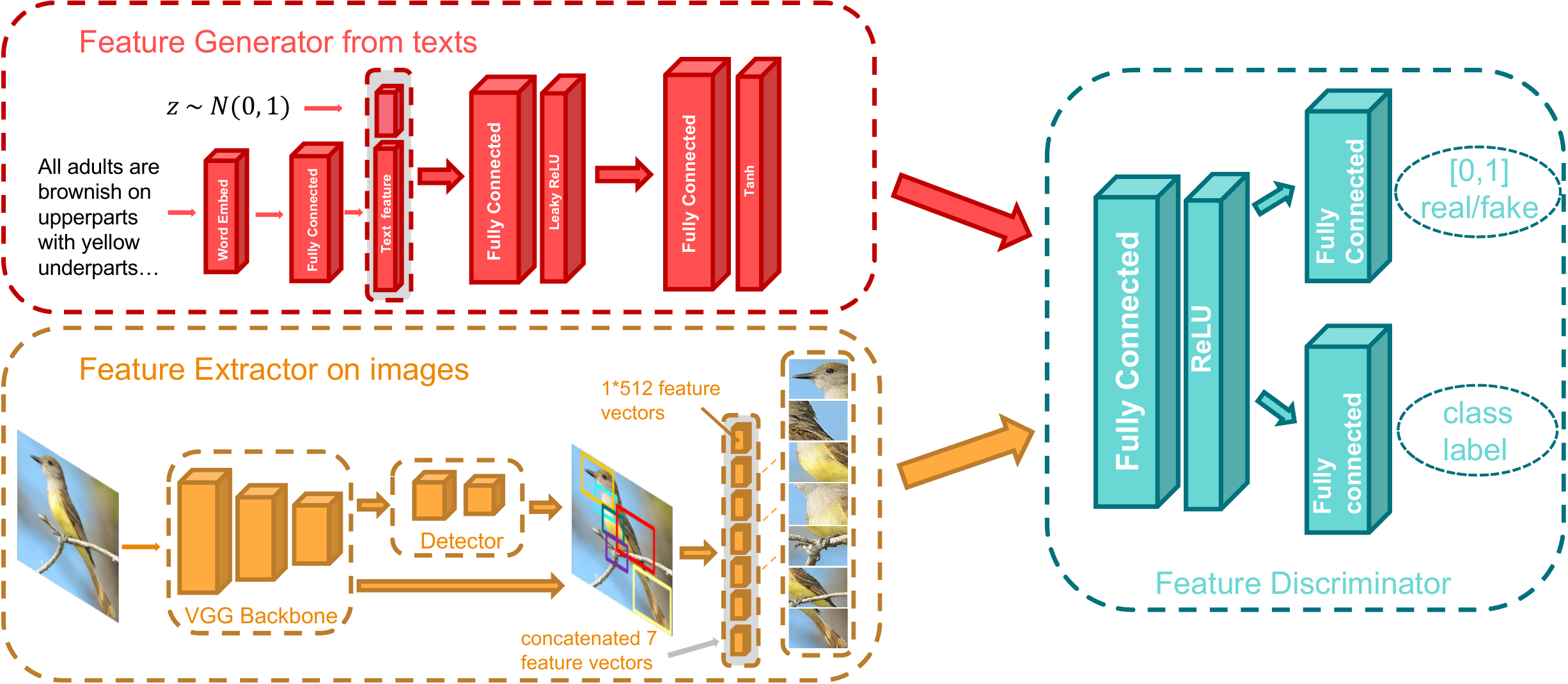}
	\caption{Model overview. Our approach first extracts deep visual features from VPDE-net (the orange part). The extracted visual features are used as the real samples for GAN. In the generator (the red part), noisy texts are embedded to semantic feature vectors of classes, then passed though a FC layer to suppress noise and reduce the dimensionality. The generator takes as input the compressed textual feature concatenated with random vector $z$ and produces the synthesized visual feature. The discriminator(the cyan part) is designed to distinguish the real or fake features and categorizes features to correct classes.}
	\label{fig:framework}
    \vspace{-5mm}
\end{figure*}

\section{Background}
In this section, we briefly describe several previous works that our method is built upon.

\subsection{Generative Adversarial Models} 
Generative adversarial networks(GANs)~\cite{goodfellow2014generative} have shown promising performance on generating realistic images~\cite{He_dehaze_2018, zhang2018txt2img, zhang2018cardiac, Tao18attngan, peng2018jointly}. They consist of a generator G and a discriminator D that contest against each other: the discriminator tries to distinguish the synthetic data from the real data while the generator tries to fool the discriminator. 
Much work has been \cite{zhao2016energy,mao2016multi, li2017mmd, arjovsky2017wasserstein, gulrajani2017improved} proposed to improve GANs by stabilizing training behavior and eliminating mode collapse, via using alternative objective losses. WGAN~\cite{arjovsky2017wasserstein} leveraged the Wasserstein distance between two distributions as the objectives, and demonstrated its capability of extinguishing mode collapse. They apply weight clipping on the discriminator to satisfy the Lipschitz constraint. The following work~\cite{gulrajani2017improved} used additional gradient penalty to replace weight clipping to get rid of the pathological behavior in ~\cite{arjovsky2017wasserstein}. 

To involve more side information to guide training procedure, conditional GANs were proposed to condition the generator and discriminator on some extra information, such as class labels~\cite{mirza2014conditional}, texts~\cite{reed2016generative, zhang2016stackgan} or even images~\cite{ isola2016image, pathak2016context, zhang2017image}.
Auxiliary Classifier GAN~\cite{odena2016conditional} further stabilized training by adding an extra category recognition branch to the discriminator. The proposed approach employed ACGAN as the basic structure while adopting the Weseertain distance with gradient penelty~\cite{gulrajani2017improved} for objectives.

\subsection{Visual Part Detector/Encoder Network}
Instead of CNN-representation of the whole image, Visual Part Detector/Encoder network (VPDE-net)~\cite{zhang2016spda} leverages the features of several semantic parts of objects for object recognition. The visual part detector has demonstrated a superior performance on visual semantic part detection. The visual part encoding proceeds by feed-forwarding the images through the conventional network (VGG as backbone) and extracting the visual features from parts detected from Visual Part Detector via ROI pooling. The encoding features of each visual part are concatenated as the visual representation of images. Our approach employs the VPDE-net as our feature extractor of images.

\section{Methodology}

We start by introducing some notations and the problem definition. The semantic representations of seen classes and unseen classes $z_i^s$ and $z_i^u$ are defined in the semantic space $\mathcal{Z}$. 
Assume $N^s$ labeled instances of seen classes $D^s = \{(x_i^s,z_i^s, y_i^s) \}^{N^s}_{i=1}$ are given as training data, where $x_i^s \in \mathcal{X} $ denotes the visual feature, $y_i^s$ is the corresponding class label. Given the visual feature $x_i^u$ of a new instance 
and a set of semantic representation of unseen classes $\{z_i^u\}_{i=1}^{N^u}$, the goal of zero-shot learning is to predict the class label $y^u$. Note that the seen class set $\mathcal{S}$ and the unseen class set $\mathcal{U}$ are disjointed, $\mathcal{S} \cap \mathcal{U} = \emptyset$.  We denote the generator as $G$:  $\mathbb{R}^Z \times \mathbb{R}^T \to \mathbb{R}^X$, the discriminator as $D$ : $\mathbb{R}^X \to \{0,1\} \times \mathbb{L}_{cls}$, where $\mathbb{L}_{cls}$ is the set of class labels. $\theta$ and $w$ are the parameters of the generator and the discriminator, respectively. 

The core of our approach is the design of a generative model to hallucinate the qualified visual features for unseen classes that further facilitate the zero-shot learning. Fig.~\ref{fig:framework} shows an overview of our generative adversarial model. Our approach adopts the basic framework of GANs for data generation. Fed with the semantic representation from the raw textual description of a class, the generator of our approach simulates the conditional distribution of visual features for the corresponding class. We employ VPDE-net to extract features of images, which serve as real samples. The discriminator is designed to distinguish real or fake features drawn from the dataset and the generator, and identify the object categories as well. See Sec~\ref{datageneration}. Additionally, to increase the distinction of synthetic feature cross the classes, visual pivot regularization is designed as an explicit cue for the generator to simulate the conditional distribution of features. See Sec~\ref{EG}. Once the pseudo features are available for each unseen class, we naturally convert zero-shot learning to a supervised classification problem. See Sec~\ref{prediction}. 


\subsection{Data Generation}
\label{datageneration}
\subsubsection{Model Achitecture} 
Our model mainly consists of three components:  Generator G to produce synthetic features; Feature extractor E to provide the real image features; Discriminator D to distinguish fake features from real ones. 


\textbf{Generator $G$}: we first embed the noisy text description using text encoder $\phi$. The text embedding $\phi(T_c)$ of class $c$ is first passed through a fully connected (FC) layer to reduce the dimensionality. We will show that this additional FC layer has a critical contribution to noise suppression. The compressed text embedding is concatenated to a random vector $z \in \mathbb{R}^Z$ sampled from Gaussian distribution $\mathcal{N} (0,1)$. The following inference proceeds by feeding it forward through two FC layers associated with two activators - Leaky Rectified Linear Unit (Leaky ReLU) and Tanh, respectively. The plausible image feature $\tilde{x}$ is generated via $\tilde{x}_c \leftarrow G_{\theta}(T_c, z) $. Feature generation corresponds to the feed-forward inference in the generator G conditioned on the text description of class $c$. The loss of generator is defined as:
\begin{equation}
L_G = - \mathbb{E}_{z \sim {p_{z}}}[D_w(G_{\theta}(T, z))] + L_{cls}(G_{\theta}(T, z)),
\label{eq:gen}
\end{equation}
where the first term is Wasserstein loss~\cite{arjovsky2017wasserstein} and the second term is the additional classification loss corresponding to class labels. 

\textbf{Discriminator $D$}: $D$ takes as input the real image features from $E$ or synthesized features from $G$, and forward them through a FC layer with ReLU.  Following this, two branches of the network are designed: (i) one FC layer for a binary classifier to distinguish if the input features are real or fake. (ii) another FC for n-ways classifier to categorize the input samples to correct classes. 
The loss for the discriminator is defined as: 
\begin{equation}
\begin{aligned}
L_D = & \mathbb{E}_{z \sim {p_{z}}}[D_w(G_{\theta}(T, z))] - \mathbb{E}_{x \sim {p_{data}}}[D_w(x)] + \lambda L_{GP} \\
&+ \frac{1}{2}(L_{cls}(G_{\theta}(T, z)) + L_{cls}(x)),
\end{aligned}
\label{eq:disc}
\end{equation}
where the first two terms approximate Wasserstein distance of the distribution of real features and fake features, the third term is the gradient penalty to enforce the Lipschitz constraint: $L_{GP} = \lambda (||\bigtriangledown_{\hat{x}} D_w(\hat{x})||_2 - 1)^2$ with $\hat{x}$ being the linear interpolation of the real feature $x$ and the fake feature $\tilde{x}$. We refer readers to~\cite{gulrajani2017improved} for more details.  The last two terms are classification losses of real and synthesized features corresponding to category labels. 

\textbf{Feature Extractor $E$}:
Following the small part proposal method proposed in~\cite{zhang2016spda}, we adopt fast-RCNN framework~\cite{girshick2015fast} with VGG16 architecture~\cite{simonyan2014very} as the backbone to detect seven semantic parts of birds. We feed forward the input image through VGG16 convolutional layers. The region proposals by~\cite{zhang2016spda} are passed through ROI pooling layer and fed into an n-way classifier (n is the number of semantic parts plus background) and a bounding box regressor. The proposed region of part $p$ with the highest confidence is considered as detected semantic part $p$. If the highest confidence is below a threshold, the part is treated as missing. The detected part regions are then fed to Visual Encoder subnetwork, where they pass through ROI pooling layer and are encoded to $512D$ feature vectors by the following FC layer.  We concatenate the feature vectors of each visual part as the visual representation of images. 
 
\subsubsection{Visual Pivot Regularization}
\label{EG}
Although the basic architecture provides a way to generating samples with the similar distribution of real visual features, it is still hard to achieve superior simulation. The potential reason is the sparsity of training samples ($\sim60$ images per class in CUB) which makes it hard to learn the distribution of high dimensional of visual feature ($\sim 3500D$). We observe the visual features of seen classes have a higher intra-class similarity and relatively lower inter-class similarity. See Fig.~\ref{fig:tSNE}.a. The distribution of visual features clearly preserves the cluster-structure in $\mathcal{X}$ space with less overlap. Motivated by this observation, we want the generated features of each class to be distributed around if not inside the corresponding cluster. To achieve it, we designed a visual pivot regularization (VP) to encourage the generator to generate features of each class that statistically match real features of that class.   

The visual pivot of each class is defined as the centroid of the cluster of visual features in $\mathcal{X}$ space. It can be either the mean calculated by averaging real visual features or the mode computed via the Mean-shift technique~\cite{comaniciu2002mean}. In practice, we find there is no difference in the performance of our approach. For simplicity, we adopt the former way, and the visual pivot corresponds to the first order moment of visual features. To be more specific, we regularize the mean of generated features of each class to be the mean of real feature distribution. The regularizer is formulated as:
\begin{equation}
\begin{aligned}
L_e = \frac{1}{C}\sum_{c=1}^{C}|| \mathbb{E}_{\tilde{x}_c \sim {p^c_{g}}}[\tilde{x}_c] - \mathbb{E}_{x_c \sim {p^c_{data}}}[x_c]||^2,\\
\end{aligned}
\label{eq：vp}
\end{equation}
where $C$ is the number of seen classes, $x_c$ is the visual feature of class $c$, and $\tilde{x}_c$ is the generated feature of class $c$, $p_g^c$ and $p^c_{data}$ are conditional distributions of synthetic and real features respectively. Since we have no access to the real distribution, in practice, we instead use the empirical expectation $\mathbb{E}_{x_c \sim {p^c_{g}}}[\hat{x}_c] = \frac{1}{N_c}\sum^{N_c}_{i=1}x^i_c$, where $N_c$ is the number of samples of class $c$ in the dataset. Similarly, the expectation of synthesized features is approximated by averaging the synthesized visual features for class $c$, $\mathbb{E}_{x_c \sim {p^c_{g}}}[\tilde{x}_c] = \frac{1}{N_c^s} \sum^{N_c^s}_{i=1}G_{\theta}(T_c, z_i)$, where $N_c^s$ is the number of synthesized features for class $c$. A technique related to our VP regularizer is feature matching proposed in ~\cite{salimans2016improved}, which aims to match statistics in the discriminator's intermediate activations w.r.t. the data distribution. 
Note that zero-shot learning is a recognition problem that favors features preserving large intra-class distinction. Compared with feature matching, matching the statistics of the data distribution can explicitly make the generator produce more distinctive features across classes.

\begin{algorithm}[t]
	\begin{algorithmic}[1]
		\STATE{\bfseries Input:} the maximal loops $N_{step}$, the batch size $m$, the iteration number of discriminator in a loop $n_{d}$, the balancing parameter $\lambda_p$, the visual pivots $\{\bar{x}_c\}^C_{c=1}$, Adam hyperparameters $\alpha$, $\beta_1$, $\beta_2$. 
        \FOR{iter $= 1,..., N_{step}$}
		\FOR{$t = 1$, ..., $n_{d}$}
        \STATE Sample a minibatch of images $x$,  matching texts $T$, random noise $z$
		\STATE $\tilde{x}   \gets G_{\theta}(T, z)$
		\STATE Compute the discriminator loss $L_D$ using Eq.~\ref{eq:disc}
        \STATE $w \gets \text{Adam}(\bigtriangledown_{w} L_D, w, \alpha, \beta_1, \beta_2)$
        \ENDFOR
        
		\STATE Initialize each set in $\{Set_c\}^C_{c=1}$ to $\varnothing$ 
        \STATE Sample a minibatch of class labels $c$, matching texts $T_c$, random noise $z$
        \STATE $\tilde{x} \gets G_{\theta}(T_c, z)$
        \STATE Compute the generator loss $L_G$ using Eq.~\ref{eq:gen}
		\STATE Add  $\tilde{x}$  to the corresponding sets of $\{Set_c\}^C_{c=1}$
		
		
		\FOR{$c =1$,..., $C$}
		\STATE$L^{(c)}_{reg} = ||mean(Set_c) - \bar{x}_c||_2$
		\ENDFOR
		
		\STATE\begin{varwidth}[t]{\linewidth}  $\theta \gets \text{Adam}(\bigtriangledown_{\theta} [ L_G + \lambda_{p}\frac{1}{C}\sum_{C}^{i=1}L^{(c)}_{reg}],	\theta, \alpha, \beta_1, \beta_2)$
		\end{varwidth}
		\ENDFOR
	\end{algorithmic}
	\caption{ Training procedure of our approach. We use default values of $n_{d} =5$, $\alpha = 0.001$, $\beta_1 = 0.5$, $\beta_2 =0.9$}
	\label{alg_label}
	
\end{algorithm}

\subsubsection{Training Procedure}
To train our model, we view visual-semantic feature pairs as joint observation. Visual features are either extracted from a feature extractor or synthesized by a generator. We train the discriminator to judge features as real or fake and predict the class labels of images, as well as optimize the generator to fool the discriminator. Algorithm~\ref{alg_label} summarizes the training procedure. In each iteration, the discriminator is optimized for $n_{d}$ steps (lines $3-7$), and the generator is optimized for $1$ step (lines $9-17$). To compute VP loss, we create $N_c$ empty sets and add each synthetic feature $\tilde{x}$ to the corresponding set w.r.t the class label (line 13).  The loss of each class is Euclidean distance between the mean of synthesized features and visual pivots (line 15).

\subsection{Zero-Shot Recognition}
\label{prediction}
With the well-trained generative model, the visual features of unseen classes can be easily synthesized by the generator with  the corresponding semantic representation.   

\begin{equation}
x_u = G_{\theta}(T_u, z)
\end{equation}

It is worth mentioning that we can generate an arbitrary number of visual features since $z$ can be sampled indefinitely. With synthesized data of unseen classes, the zero-shot recognition becomes a conventional classification problem. In practice, any supervised classification methods can be employed. In this paper, we simply use nearest neighbor prediction to demonstrate the ability of our generator.

\section{Experiments}
\subsection{Experiment Setting}
\noindent \textbf{Datasets:} We evaluated our method with state-of-the-art
approaches on two benchmark datasets: \emph{Caltech UCSD Birds-2011} (CUB)~\cite{wah2011caltech} and \emph{North America Birds} (NAB)~\cite{Horn2015}. Both are datasets of birds for fine-grained classification. The CUB dataset contains
200 categories of bird species with a total of 11,788
images, and NAB is a larger dataset of birds with 1011 classes and 48,562 images. Elhoseiny \emph{et al.}~\cite{Elhoseiny_2017_CVPR} extended both datasets by adding the Wikipedia article of each class, and they also reorganized NAB to 404 classes by merging subtle
division of classes, such as ``American Kestrel
(Female, immature)" and ``American Kestrel (Adult male)". 
In ~\cite{Elhoseiny_2017_CVPR}, two different split settings were proposed for both datasets, named \emph{Super-Category-Shared} and \emph{Super-Category-Exclusive} splittings, in term of how close the seen classes are related to the unseen ones. For brevity, we denote them as SCS-split and SCE-split. In the scenario of SCS-split, for each unseen class, there exists one or more seen classes that belong to the same parent category. For instance, both ``Cooper's
Hawk" in the training set and ``Harris's Hawk" in the testing set are under the parent category ``Hawks". Note that the conventional ZSL setting is SCS-split used in ~\cite{akata2016multi, Qiao2016, romera2015embarrassingly, akata2015evaluation}. On the contrary, in SCS-split, the parent categories of unseen classes are exclusive to those of the seen classes. Intuitively, SCE-split is much harder than SCS-split as the relevance between seen and unseen classes is minimized. We follow both split settings to evaluate the capability of our approach.

\noindent \textbf{Textual Representation:}
We use the raw Wikipedia articles collected by~\cite{Elhoseiny_2017_CVPR} for both benchmark datasets.
Text articles are first tokenized into
words, the stop words are removed, and porter stemmor~\cite{Porter} is applied to reduce inflected words to their word stem.  Then, Term
Frequency-Inverse Document Frequency(TF-IDF) feature
vector~\cite{salton1988term} is extracted. The dimensionalities of TF-IDF features for CUB and
NAB are 7551 and 13217.

\noindent \textbf{Visual Features:}
We extract the visual features from the activations of the part-based FC layer of VPDE-net. All input images are resized to $224 \times 224$ and fed into the VPDE-net. There are seven semantic parts in CUB dataset: ``head", ``back", ``belly", ``breast", ``leg", ``wing", ``tail". NAB dataset contains the same semantic parts except for ``leg". For each part, a 512-dimensional feature vector is extracted. Concatenating those part-based features in order, we obtain the visual representation of the image. The dimensionalities of visual features for CUB and NAB are 3583 and 3072 respectively. 

The implementation details of our model and the parameter settings can be found in the supplementary material. 

\subsection{Zero-Shot Recognition}
\noindent \textbf{Competing Methods:}
The performance of our method is compared to seven state-of-the-art algorithms: ZSLPP~\cite{Elhoseiny_2017_CVPR}, MCZSL~\cite{akata2016multi}, ZSLNS~\cite{Qiao2016}, ESZSL~\cite{romera2015embarrassingly}, SJE~\cite{akata2015evaluation}, WAC~\cite{elhoseiny2013write}, SynC~\cite{changpinyo2016synthesized}.
The source code of ZSLPP, ESZSL, and ZSLNS are available online,
and we get the code of WAC~\cite{elhoseiny2013write} from its author. For MCZSL and SJE, since their source codes are not available,
we directly copy the highest scores for non-attribute settings
reported in~\cite{akata2016multi, akata2015evaluation}.


We conduct the experiments on both SCS and SCE splits on two benchmark datasets to show the performance of our approach.   Note that some of the compared methods are attribute-based methods but applicable in our setting by replacing the attribute vectors with textual features.  Among these methods MCZSL and ZSLPP leverage the semantic parts of birds for visual representations of images. MCZSL directly uses part annotations as strong supervision to extract CNN representation of each semantic part in the test phase. Unlike MCZSL, our approach and ZSLPP are merely based on the detected semantic parts during both training and testing. The performance of the final zero-shot classification is expected to degrade due to less accurate detection of semantic parts compared to manual annotation in MCZSL. Table~\ref{tb:cub2011} shows the performance comparisons on CUB and NAB datasets. Generally, our method consistently outperforms the state-of-the-art methods. On the conventional split setting (SCS), our approach outperforms the runner-up (ZSLPP) by a considerable gap: $6.5\%$ and $5.3\%$ on CUB dataset and NAB dataset, respectively. Note that ZSL on SCE-split remains rather challenging. The fact that there is less relevant information between the training and testing set makes it hard to transfer knowledge from seen classes to unseen classes. Although our method just improves the performance by less than 1\%, we will show the great improvement on the general merit of ZSL in Sec~\ref{sec:gzsl}  

\setlength{\tabcolsep}{3pt}
\begin{table}[h]
	\centering 
	\scalebox{0.8}
	{
		\begin{tabular}{l| K{1.2cm}| K{1.2cm} | K{1.2cm}| K{1.2cm}}
			\Xhline{4\arrayrulewidth}
			  &\multicolumn{2}{c|}{CUB}  &\multicolumn{2}{c}{NAB}  \\ 
			\hline
			methods &SCS &SCE   &SCS  &SCE  \\ 
	\Xhline{4\arrayrulewidth}
			&&&&\\[-0.8em]
			MCZSL \cite{akata2016multi}   	&  34.7 & --  &--&--\\
			WAC-Linear \cite{elhoseiny2013write}  	&  27.0  &   5.0 &--&--\\	
			WAC-Kernel \cite{elhoseiny2016write}    	&  33.5    & 7.7 & 11.4 &6.0\\	
			ESZSL \cite{romera2015embarrassingly}     	&  28.5   &7.4 &24.3&6.3\\	
			SJE \cite{akata2015evaluation}  & 29.9 & --&--&--\\
			ZSLNS \cite{Qiao2016}  	&  29.1   &7.3 &24.5&6.8\\	
			SynC$_{fast}$ \cite{changpinyo2016synthesized} & 28.0 & 8.6 &18.4&3.8\\
			SynC$_{OVO}$ \cite{changpinyo2016synthesized} & 12.5 &  5.9 &--&--\\
			ZSLPP \cite{Elhoseiny_2017_CVPR}    	&  37.2  &9.7 &30.3&8.1\\
			GAZSL   	&  \textbf{43.7}  &\textbf{10.3} &\textbf{35.6}&\textbf{8.6}\\
			\Xhline{4\arrayrulewidth}
			
		\end{tabular}
	}
	\vspace{1mm}
	\caption{
		{ Top-1 accuracy (\%) on \textbf{CUB} and \textbf{NAB} datasets with two split settings.}} 
	\vspace{-7.5mm}
	\label{tb:cub2011}
\end{table}

\subsubsection{Ablation Study}
We now do the ablation study of the effect of the visual pivot regularization(VP) and the GAN. We trained two variants of our model by only keeping the VP or GAN, denoted as VP-only and GAN-only, respectively. Specifically, in the case of only using VP, our model discards the discriminator and thus is reduced to a visual semantic embedding method with VP loss. The generator essentially becomes a mapping function that projects the semantic feature of classes to the visual feature space. It is worth mentioning that compared to previous linear embedding methods~\cite{elhoseiny2013write, romera2015embarrassingly,Qiao2016,Elhoseiny_2017_CVPR} that usually have one or two projection matrices, our generator has a deeper architecture with three projection matrices if we can roughly treat FC layers as project matrices. Additionally, our generator also differs in adding nonlinearity by using Leaky-ReLU and Tanh as the activation function as shown in Fig.~\ref{fig:framework}.

Table~\ref{tb:ablation} shows the performance of each setting. Without the visual pivot regularization, the performance drops drastically by 20.91\% (22.83\%vs.43.74\%) on CUB and 11.36\%     
(24.22\%vs.35.58\%) on NAB, highlighting the importance of the designed VP regularizer to provide a proper explicit supervision to GAN. 
Interestingly, we observe that only using VP regularizer as the objective of the generator achieves the accuracy of 28.52\% on CUB and 25.75\% on NAB, which are even higher than that of GAN-only model. This observation naturally introduces another perspective of our approach. We can regard our approach as a visual semantic embedding one with the GAN as a constraint. We argue that generative adversarial model and the visual pivot regularization are critically complementary to each other. GAN makes it possible to get rid of the one-to-one mapping existing in previous embedding methods by generating diverse embeddings on descriptions.
On the other hand, VP regularization attempts to restrict such imaginary samples of the generator within the proper range, thus keeping the discrimination of generated features across the classes. The importance of both components is verified by the superior performance of our complete approach compared to two variants. 
\begin{table}[h]
	
	\centering 
	\scalebox{0.8}
	{
		\begin{tabular}{l|K{1.3cm}|K{1.3cm}|K{1.3cm}|K{1.3cm}}
					
			\Xhline{4\arrayrulewidth}
			&\multicolumn{2}{c|}{CUB}  &\multicolumn{2}{c}{NAB} \\\hline
			&&&&\\[-0.8em]
			methods & w/ FC  &w/o FC & w/ FC  &w/o FC\\ 
			\Xhline{4\arrayrulewidth}
			&&&&\\[-0.8em]
			
			GAN-only &  22.83   &21.83 &24.22& 24.80\\
			VP-only  &28.52 &26.76 &25.75&23.42\\
			GAZSL	&  \textbf{43.74}   &40.85 &\textbf{35.58}&32.94\\
			\Xhline{4\arrayrulewidth}	
		\end{tabular}
	}
	\caption{
		Effects of different components on zero-shot classification accuracy (\%) on CUB and NAB datasets with SCS split setting.
	}
	\vspace{-3mm}
	\label{tb:ablation}
\end{table}

We also analyzed the effectiveness of our additional FC layer on textual features for noise suppression. As shown in Table~\ref{tb:ablation}, in general, our method with FC layer outperforms one without FC layer by $2\%$ to $3\%$ in most cases. The superiority can also be observed in two variants. In practice, the high dimensional TF-IDF feature is compacted to a $1000D$ feature vector. Unlike the traditional dimensionality  reduction technique (e.g., PCA), FC layer contains trainable weights and is optimized in an end-to-end fashion.

\begin{figure}
	\centering
	\subfloat[CUB with SCS splitting]{\includegraphics[width= 1.6in]{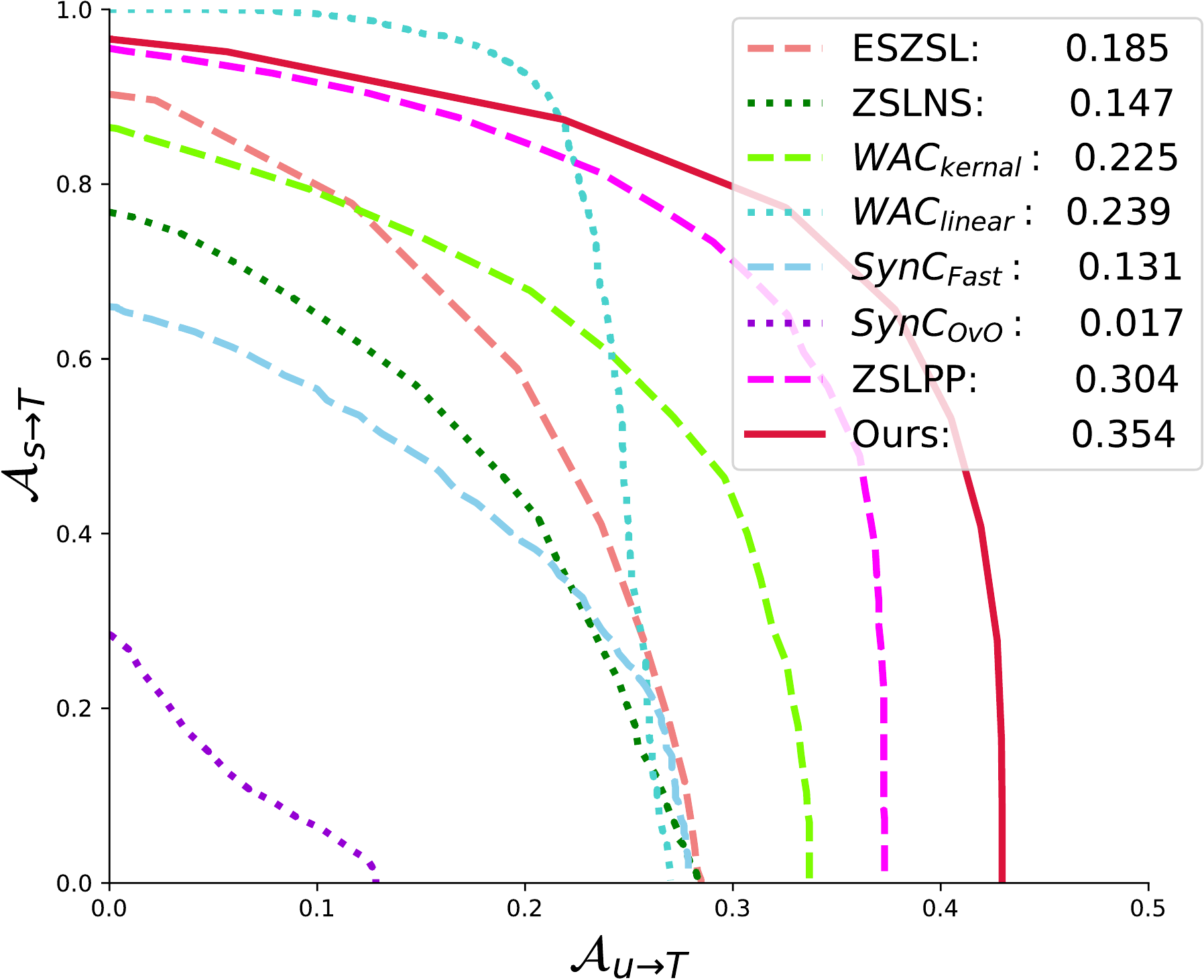}} 
	\subfloat[CUB with SCE splitting]{\includegraphics[width= 1.6in]{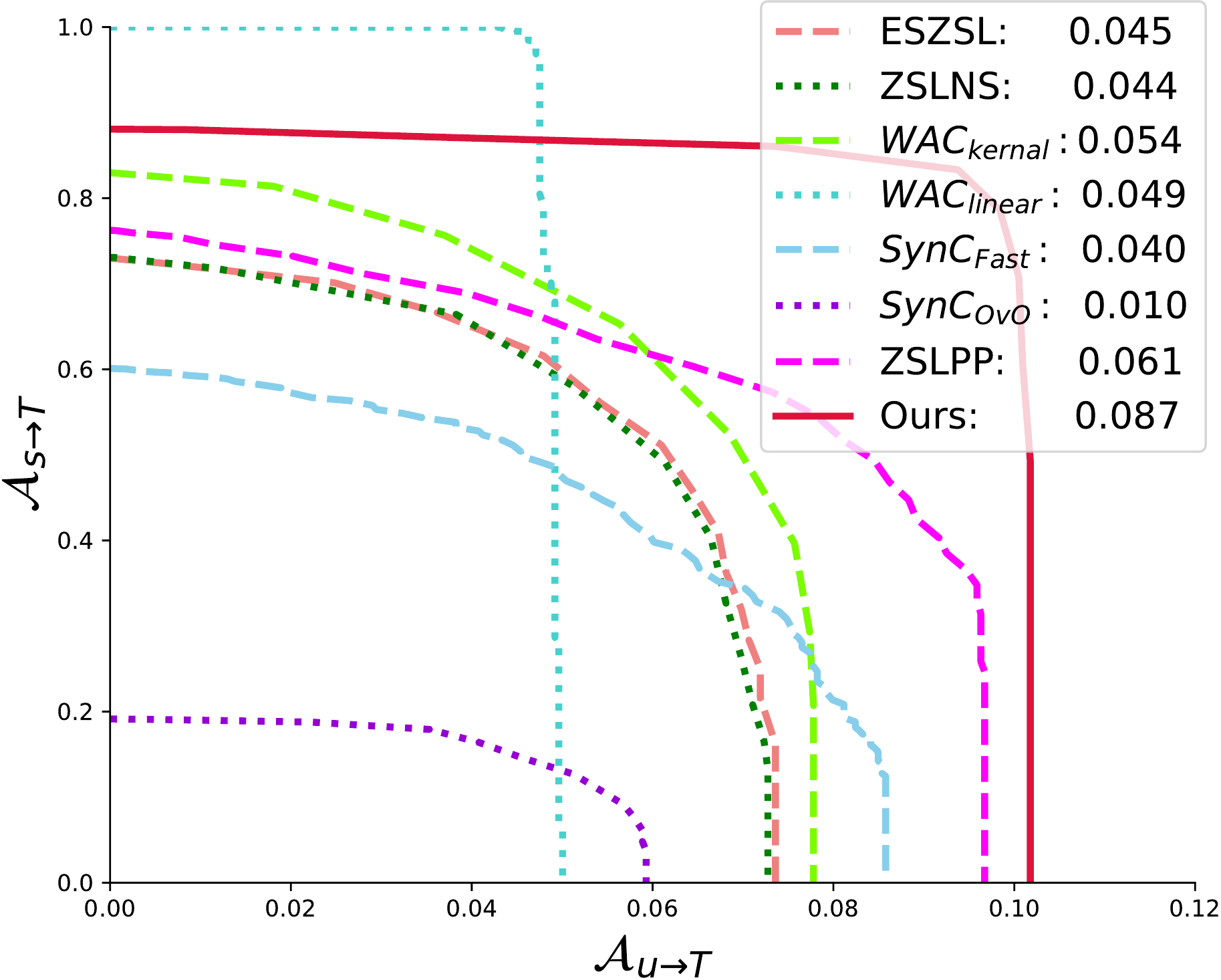}}\\
    \vspace{-2mm}
	\subfloat[NAB with SCS splitting]{\includegraphics[width= 1.6in]{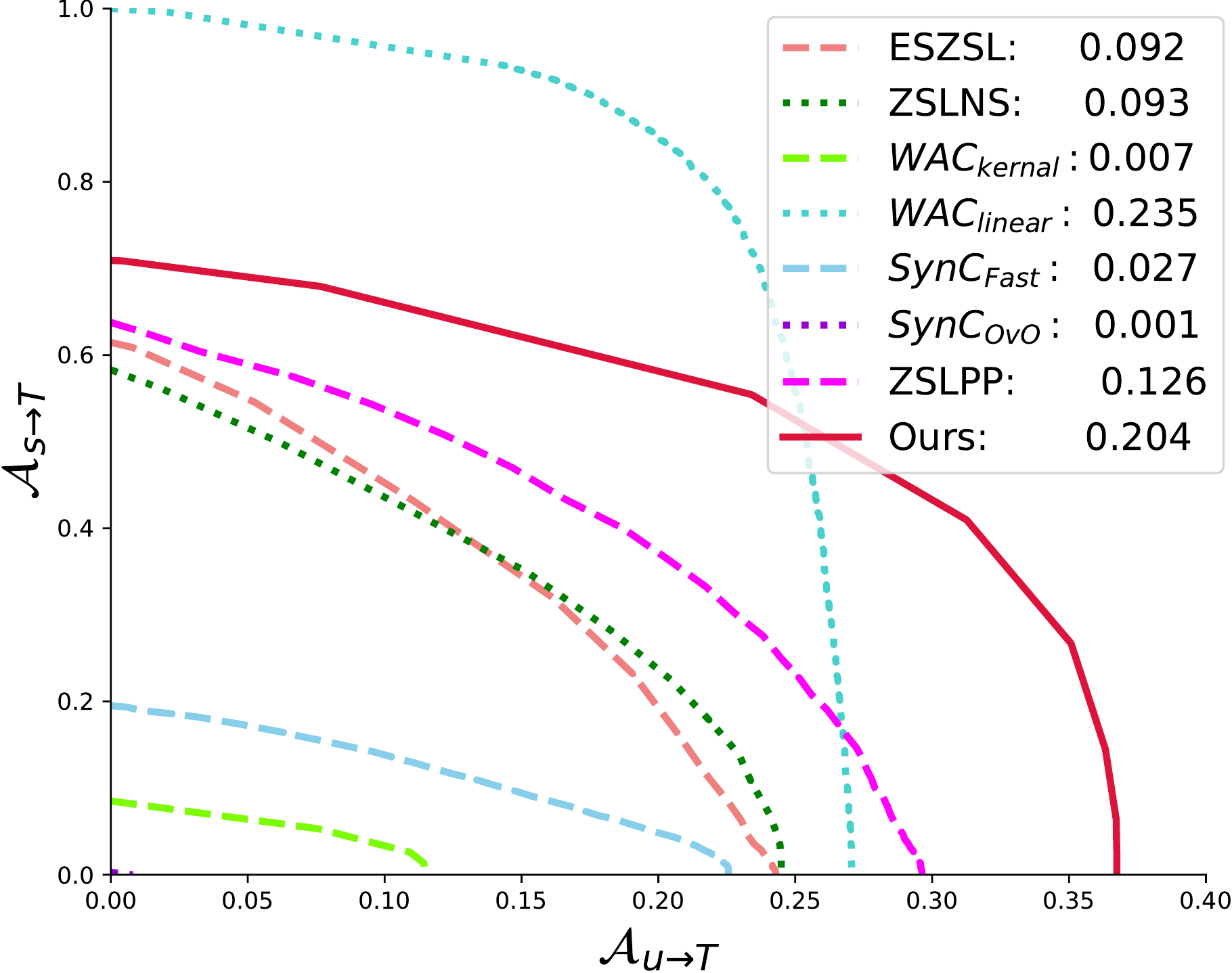}} 
	\subfloat[NAB with SCE splitting]{\includegraphics[width= 1.6in]{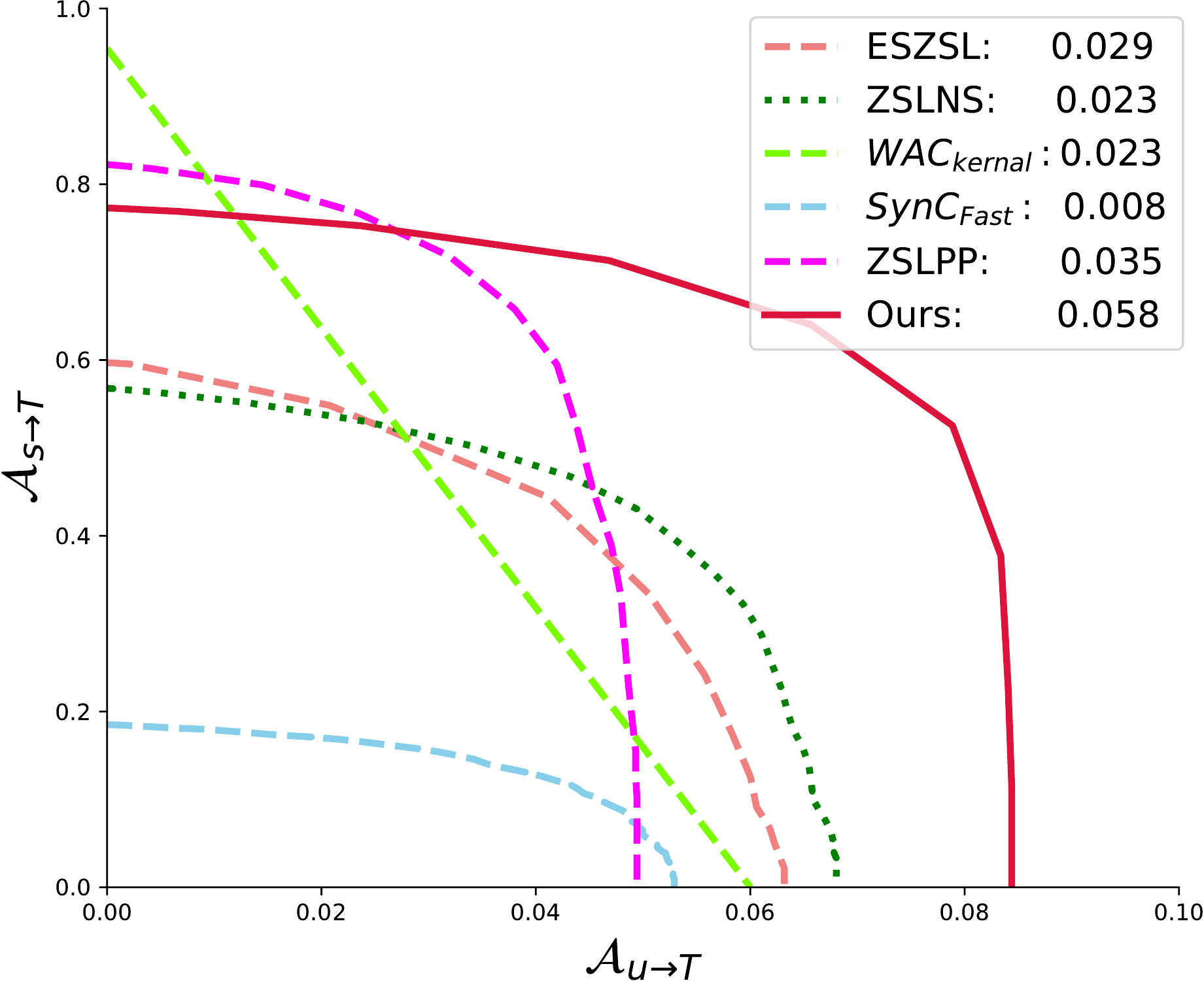}}\\
	\setlength\belowcaptionskip{-2.5ex}
	\caption{ Seen-Unseen accuracy
		Curve on two benchmarks datasets with two split settings}
	\label{fig:GZSL_Curve}
    \vspace{-4mm}
\end{figure}

\subsection{Generalized Zero-Shot Learning}
\label{sec:gzsl}
The conventional zero-shot recognition considers that queries come from only unseen classes. However, as the seen classes are often the most common objects, it is unrealistic to assume that we will never encounter them during the test phase~\cite{chao2016empirical}
. Chao \emph{et al.}~\cite{chao2016empirical} presented a more general metric for ZSL that involves classifying images of both seen classes $\mathcal{S}$ and unseen classes $\mathcal{U}$ into $\mathcal{T} = \mathcal{S}\cup\mathcal{U}$. The accuracies are denoted as $A_{\mathcal{S}\to\mathcal{T}}$ and $A_{\mathcal{U}\to\mathcal{T}}$ respectively. They introduced a balancing parameter $\lambda$ to draw Seen-Unseen accuracy Curve(SUC) and use Area Under SUC  to measure the general capability of methods for ZSL. In our case, we use the trained GAN to synthesize the visual features of both training classes and testing classes. The visual features of each class are averaged to obtain the visual pivots. 
The nearest neighbor strategy to visual pivots is adopted to predict the class label of images. 

In Fig.~\ref{fig:GZSL_Curve}, we plot SUC and report the AUSUC scores of our method and the competitors. Our method compares favorably to competitors in all cases except on NAB with SCS-split, where very high $A_{\mathcal{S}\to\mathcal{T}}$ and low $A_{\mathcal{U}\to\mathcal{T}}$ indicate that $WAC_{linear}$ is overfitting the dataset. It's worth noting that although our method only slightly outperforms competitors in the zero-shot recognition task with SCE splitting, the AUSUC scores of our method are $42.6\%$ and $56.7\%$ higher than those of the runner-up on CUB and NAB respectively, indicating a superior and balanced performance on $A_{\mathcal{S}\to\mathcal{T}}$ and $A_{\mathcal{U}\to\mathcal{T}}$.

\subsection{Zero-Shot Retrieval}
Zero-shot image retrieval is defined as retrieving images giving the semantic representation of an unseen class as the query. We use mean average precision (mAP) to evaluate the performance. In Table~\ref{tb:Retrieval}, we report the performance of different settings: retrieving 25\%, 50\%, 100\% of the number of images for each class from the whole dataset. The precision is defined as the ratio of the number of correct retrieved images to that of all retrieved images.  We adopt the same strategy as in GZSL to obtain the visual pivots of unseen classes. Given the visual pivot, we retrieve images based on the nearest neighbor strategy in $\mathcal{X}$ space.

\begin{table}[h]
	\centering 
	\scalebox{0.82}
	{
		\begin{tabular}{l|K{0.9cm}|K{0.9cm}|K{0.9cm}|K{0.9cm}|K{0.9cm}|K{0.9cm}}
			\Xhline{4\arrayrulewidth}
			 &\multicolumn{3}{c|}{CUB}  &\multicolumn{3}{c}{NAB} \\\hline
			&&&&&&\\[-0.8em]
			methods &25\%  &50\%  &100\%   &25\%  &50\%  &100\%\\ 
			\Xhline{4\arrayrulewidth}
			&&&&&&\\[-0.8em]
			ESZSL \cite{romera2015embarrassingly} 	&   27.9   &27.3       &22.7 &28.9&27.8&20.85\\
			
			ZSLNS \cite{Qiao2016}	&  29.2    &29.5      &23.9&28.78&27.27&22.13\\
			
		ZSLPP \cite{Elhoseiny_2017_CVPR}	&  42.3    &42.0       &36.3&36.9&35.7&\textbf{31.3}\\
			
			VP-only&  17.8   &16.4  & 13.9&15.1&13.1&11.5\\
			GAN-only	&  18.0   &17.5  & 15.2&21.7&20.3&16.6\\
			GAZSL	&  \textbf{49.7 }  &\textbf{48.3} & \textbf{40.3}&\textbf{41.6}&\textbf{37.8}&31.0\\
			\Xhline{4\arrayrulewidth}	
		\end{tabular}
	}
	\caption{Zero-Shot Retrieval using mean Average Precision
(mAP) (\%) on CUB and NAB with SCS splitting.
	}
	\vspace{-3mm}
	\label{tb:Retrieval}
\end{table}
Overall, our method outperforms competitors by $4\% \sim 7\%$. 
We discover that although VP-only performs better than GAN-only on the recognition task, its performance is inferior to GAN-only on retrieval tasks. Without the  constraints of WGAN, VP-only suffers from heavy mode collapse, and synthetic features easily collapse into the visual pivot of each class with less diversity.   

\begin{figure}
\centering
	\includegraphics[width=7.8cm,height=6.5cm]{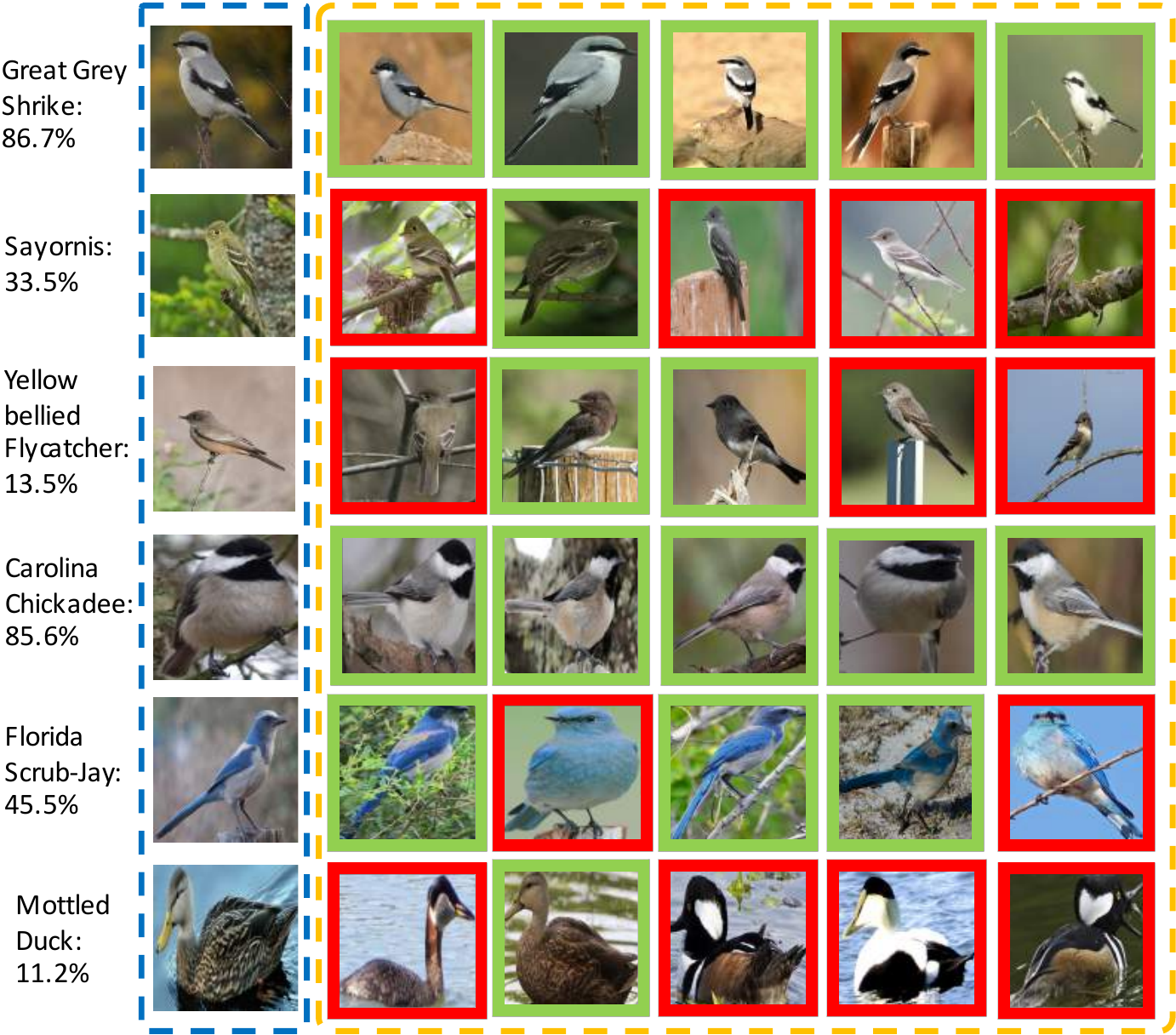}
	\caption{Qualitative results of zero-shot retrieval. The first three rows are classes from CUB and the rest from NAB. Correct and incorrect retrieved instances are shown in green and red respectively.}
	\label{fig:retivial}
	\vspace{-5mm}
\end{figure}

We also provide qualitative results of our method as shown in Fig.~\ref{fig:retivial}. Each row is one class, and the class name and the precision are shown on the left. The first column is Top-1 within-class nearest neighbor. The following five columns are Top-5 overall nearest neighbors without considering the instances in the first column. 

\subsection{t-SNE Demonstration}
Fig.~\ref{fig:tSNE} demonstrates the t-SNE~\cite{maaten2008visualizing} visualization of the real features, the synthesized features for unseen classes under different settings. Despite some overlaps, the real features roughly distribute in separate clusters w.r.t class labels. The features generated by our method keep the same structure. Ideally, we expect the distribution of generated features matches that of the real ones in $\mathcal{X}$ space.  Empirically, despite that some deviating clusters of synthesized features  due to the bias between testing and training data, many clusters are well aligned to the real ones. Without the VP regularizer, GAN encourages the high diversity of features with no constraints. The features disperse greatly and lack the discrimination across classes.

\section{Conclusion}
In this paper, we proposed a novel generative adversarial approach for ZSL, which leverages GANs to dispersively imagine the visual features given the noisy textual descriptions from Wikipedia. 
We introduced the visual pivot regularizer to explicitly guide the imagery samples of GANs to the proper direction. We also showed that adding a FC layer for textual feature results in comparable noise suppression.
Experiments showed that our approach consistently performs favorably against the state-of-the-art methods on multiple zero-shot tasks, with an outstanding capability of visual feature generation. 

\textbf{ Acknowledgment. } This work was supported NSF-USA award \#1409683.

\clearpage

{\small
\bibliographystyle{ieee}
\bibliography{egbib}

\begin{thebibliography}{10}\itemsep=-1pt

\bibitem{akata2016multi}
Z.~Akata, M.~Malinowski, M.~Fritz, and B.~Schiele.
\newblock Multi-cue zero-shot learning with strong supervision.
\newblock In {\em {CVPR}}, 2016.

\bibitem{akata2016label}
Z.~Akata, F.~Perronnin, Z.~Harchaoui, and C.~Schmid.
\newblock Label-embedding for image classification.
\newblock {\em IEEE Transactions on Pattern Analysis and Machine Intelligence},
  38(7):1425--1438, 2016.

\bibitem{akata2015evaluation}
Z.~Akata, S.~Reed, D.~Walter, H.~Lee, and B.~Schiele.
\newblock Evaluation of output embeddings for fine-grained image
  classification.
\newblock In {\em {CVPR}}, 2015.

\bibitem{arjovsky2017wasserstein}
M.~Arjovsky, S.~Chintala, and L.~Bottou.
\newblock Wasserstein gan.
\newblock {\em arXiv preprint arXiv:1701.07875}, 2017.

\bibitem{changpinyo2016synthesized}
S.~Changpinyo, W.-L. Chao, B.~Gong, and F.~Sha.
\newblock Synthesized classifiers for zero-shot learning.
\newblock In {\em {CVPR}}, pages 5327--5336, 2016.

\bibitem{chao2016empirical}
W.-L. Chao, S.~Changpinyo, B.~Gong, and F.~Sha.
\newblock An empirical study and analysis of generalized zero-shot learning for
  object recognition in the wild.
\newblock In {\em {ECCV}}, pages 52--68. Springer, 2016.

\bibitem{comaniciu2002mean}
D.~Comaniciu and P.~Meer.
\newblock Mean shift: A robust approach toward feature space analysis.
\newblock {\em IEEE Transactions on Pattern Analysis and Machine Intelligence},
  24(5):603--619, 2002.

\bibitem{elhoseiny2016write}
M.~Elhoseiny, A.~Elgammal, and B.~Saleh.
\newblock Write a classifier: Predicting visual classifiers from unstructured
  text.
\newblock {\em IEEE Transactions on Pattern Analysis and Machine Intelligence},
  2016.

\bibitem{elhoseiny2013write}
M.~Elhoseiny, B.~Saleh, and A.~Elgammal.
\newblock Write a classifier: Zero-shot learning using purely textual
  descriptions.
\newblock In {\em {ICCV}}, 2013.

\bibitem{Elhoseiny_2017_CVPR}
M.~Elhoseiny, Y.~Zhu, H.~Zhang, and A.~Elgammal.
\newblock Link the head to the "beak": Zero shot learning from noisy text
  description at part precision.
\newblock In {\em {CVPR}}, July 2017.

\bibitem{frome2013devise}
A.~Frome, G.~S. Corrado, J.~Shlens, S.~Bengio, J.~Dean, T.~Mikolov, et~al.
\newblock Devise: A deep visual-semantic embedding model.
\newblock In {\em {NIPS}}, pages 2121--2129, 2013.

\bibitem{fu2016semi}
Y.~Fu and L.~Sigal.
\newblock Semi-supervised vocabulary-informed learning.
\newblock In {\em {CVPR}}, pages 5337--5346, 2016.

\bibitem{girshick2015fast}
R.~Girshick.
\newblock Fast r-cnn.
\newblock In {\em {ICCV}}, pages 1440--1448, 2015.

\bibitem{goodfellow2014generative}
I.~Goodfellow, J.~Pouget-Abadie, M.~Mirza, B.~Xu, D.~Warde-Farley, S.~Ozair,
  A.~Courville, and Y.~Bengio.
\newblock Generative adversarial nets.
\newblock In {\em {NIPS}}, pages 2672--2680, 2014.

\bibitem{gulrajani2017improved}
I.~Gulrajani, F.~Ahmed, M.~Arjovsky, V.~Dumoulin, and A.~Courville.
\newblock Improved training of wasserstein gans.
\newblock {\em arXiv preprint arXiv:1704.00028}, 2017.

\bibitem{guo2017synthesizing}
Y.~Guo, G.~Ding, J.~Han, and Y.~Gao.
\newblock Synthesizing samples for zero-shot learning.
\newblock In {\em {IJCAI}}, 2017.

\bibitem{guo2017zero}
Y.~Guo, G.~Ding, J.~Han, and Y.~Gao.
\newblock Zero-shot learning with transferred samples.
\newblock {\em IEEE Transactions on Image Processing}, 2017.

\bibitem{isola2016image}
P.~Isola, J.-Y. Zhu, T.~Zhou, and A.~A. Efros.
\newblock Image-to-image translation with conditional adversarial networks.
\newblock In {\em {CVPR}}, 2017.

\bibitem{jiang1997semantic}
J.~J. Jiang and D.~W. Conrath.
\newblock Semantic similarity based on corpus statistics and lexical taxonomy.
\newblock {\em Proceedings of International Conference Research on
  Computational Linguistics}, 1997.

\bibitem{lampert2009}
C.~H. Lampert, H.~Nickisch, and S.~Harmeling.
\newblock Learning to detect unseen object classes by between-class attribute
  transfer.
\newblock In {\em {CVPR}}, pages 951--958. IEEE, 2009.

\bibitem{Lampert2014}
C.~H. Lampert, H.~Nickisch, and S.~Harmeling.
\newblock Attribute-based classification for zero-shot visual object
  categorization.
\newblock {\em IEEE Transactions on Pattern Analysis and Machine Intelligence},
  36(3):453--465, March 2014.

\bibitem{lei2015predicting}
J.~Lei~Ba, K.~Swersky, S.~Fidler, et~al.
\newblock Predicting deep zero-shot convolutional neural networks using textual
  descriptions.
\newblock In {\em {ICCV}}, 2015.

\bibitem{li2017mmd}
C.-L. Li, W.-C. Chang, Y.~Cheng, Y.~Yang, and B.~P{\'o}czos.
\newblock Mmd gan: Towards deeper understanding of moment matching network.
\newblock In {\em {NIPS}}, 2017.

\bibitem{long2017zero}
Y.~Long, L.~Liu, L.~Shao, F.~Shen, G.~Ding, and J.~Han.
\newblock From zero-shot learning to conventional supervised classification:
  Unseen visual data synthesis.
\newblock In {\em {CVPR}}, 2017.

\bibitem{maaten2008visualizing}
L.~v.~d. Maaten and G.~Hinton.
\newblock Visualizing data using t-sne.
\newblock {\em Journal of Machine Learning Research}, 9(Nov):2579--2605, 2008.

\bibitem{mao2016multi}
X.~Mao, Q.~Li, H.~Xie, R.~Y. Lau, and Z.~Wang.
\newblock Multi-class generative adversarial networks with the l2 loss
  function.
\newblock {\em arXiv preprint arXiv:1611.04076}, 2016.

\bibitem{mirza2014conditional}
M.~Mirza and S.~Osindero.
\newblock Conditional generative adversarial nets.
\newblock {\em arXiv preprint arXiv:1411.1784}, 2014.

\bibitem{odena2016conditional}
A.~Odena, C.~Olah, and J.~Shlens.
\newblock Conditional image synthesis with auxiliary classifier gans.
\newblock {\em arXiv preprint arXiv:1610.09585}, 2016.

\bibitem{pathak2016context}
D.~Pathak, P.~Krahenbuhl, J.~Donahue, T.~Darrell, and A.~A. Efros.
\newblock Context encoders: Feature learning by inpainting.
\newblock In {\em {CVPR}}, pages 2536--2544, 2016.

\bibitem{peng2018jointly}
X.~Peng, Z.~Tang, F.~Yang, S.~R. Feris, and M.~Dimitris.
\newblock Jointly optimize data augmentation and network training: Adversarial
  data augmentation in human pose estimation.
\newblock In {\em {CVPR}}, 2018.

\bibitem{Porter}
M.~F. Porter.
\newblock Readings in information retrieval.
\newblock chapter An Algorithm for Suffix Stripping, pages 313--316. Morgan
  Kaufmann Publishers Inc., San Francisco, CA, USA, 1997.

\bibitem{Qiao2016}
R.~Qiao, L.~Liu, C.~Shen, and A.~v.~d. Hengel.
\newblock Less is more: Zero-shot learning from online textual documents with
  noise suppression.
\newblock In {\em {CVPR}}, June 2016.

\bibitem{reed2016learning}
S.~Reed, Z.~Akata, B.~Schiele, and H.~Lee.
\newblock Learning deep representations of fine-grained visual descriptions.
\newblock In {\em {CVPR}}, 2016.

\bibitem{reed2016generative}
S.~Reed, Z.~Akata, X.~Yan, L.~Logeswaran, B.~Schiele, and H.~Lee.
\newblock Generative adversarial text to image synthesis.
\newblock In {\em {ICML}}, 2016.

\bibitem{resnik1995using}
P.~Resnik.
\newblock Using information content to evaluate semantic similarity in a
  taxonomy.
\newblock In {\em Proceedings of the 14th International Joint Conference on
  Artificial Intelligence - Volume 1}, IJCAI'95, pages 448--453, 1995.

\bibitem{romera2015embarrassingly}
B.~Romera-Paredes and P.~Torr.
\newblock An embarrassingly simple approach to zero-shot learning.
\newblock In {\em {ICML}}, pages 2152--2161, 2015.

\bibitem{salimans2016improved}
T.~Salimans, I.~Goodfellow, W.~Zaremba, V.~Cheung, A.~Radford, and X.~Chen.
\newblock Improved techniques for training gans.
\newblock In {\em {NIPS}}, pages 2234--2242, 2016.

\bibitem{salton1988term}
G.~Salton and C.~Buckley.
\newblock Term-weighting approaches in automatic text retrieval.
\newblock {\em Information processing \& management}, 24(5):513--523, 1988.

\bibitem{shigeto2015ridge}
Y.~Shigeto, I.~Suzuki, K.~Hara, M.~Shimbo, and Y.~Matsumoto.
\newblock Ridge regression, hubness, and zero-shot learning.
\newblock In {\em Joint European Conference on Machine Learning and Knowledge
  Discovery in Databases}, pages 135--151. Springer, 2015.

\bibitem{simonyan2014very}
K.~Simonyan and A.~Zisserman.
\newblock Very deep convolutional networks for large-scale image recognition.
\newblock In {\em {ICLR}}, 2015.

\bibitem{socher2013zero}
R.~Socher, M.~Ganjoo, C.~D. Manning, and A.~Ng.
\newblock Zero-shot learning through cross-modal transfer.
\newblock In {\em {NIPS}}, pages 935--943, 2013.

\bibitem{tsai2017learning}
Y.-H.~H. Tsai, L.-K. Huang, and R.~Salakhutdinov.
\newblock Learning robust visual-semantic embeddings.
\newblock In {\em {ICCV}}, 2017.

\bibitem{Horn2015}
G.~Van~Horn, S.~Branson, R.~Farrell, S.~Haber, J.~Barry, P.~Ipeirotis,
  P.~Perona, and S.~Belongie.
\newblock Building a bird recognition app and large scale dataset with citizen
  scientists: The fine print in fine-grained dataset collection.
\newblock In {\em {CVPR}}, 2015.

\bibitem{wah2011caltech}
C.~Wah, S.~Branson, P.~Welinder, P.~Perona, and S.~Belongie.
\newblock The caltech-ucsd birds-200-2011 dataset.
\newblock 2011.

\bibitem{Tao18attngan}
T.~Xu, P.~Zhang, Q.~Huang, H.~Zhang, Z.~Gan, X.~Huang, and X.~He.
\newblock Attngan: Fine-grained text to image generation with attentional
  generative adversarial networks.
\newblock In {\em {CVPR}}, 2018.

\bibitem{yang2014unified}
Y.~Yang and T.~M. Hospedales.
\newblock A unified perspective on multi-domain and multi-task learning.
\newblock In {\em {ICLR}}, 2015.

\bibitem{He_dehaze_2018}
H.~Zhang and V.~M. Patel.
\newblock Densely connected pyramid dehazing network.
\newblock In {\em {CVPR}}, 2018.

\bibitem{zhang2017image}
H.~Zhang, V.~Sindagi, and V.~M. Patel.
\newblock Image de-raining using a conditional generative adversarial network.
\newblock {\em arXiv preprint arXiv:1701.05957}, 2017.

\bibitem{zhang2016spda}
H.~Zhang, T.~Xu, M.~Elhoseiny, X.~Huang, S.~Zhang, A.~Elgammal, and D.~Metaxas.
\newblock Spda-cnn: Unifying semantic part detection and abstraction for
  fine-grained recognition.
\newblock In {\em {CVPR}}, pages 1143--1152, 2016.

\bibitem{zhang2016stackgan}
H.~Zhang, T.~Xu, H.~Li, S.~Zhang, X.~Wang, X.~Huang, and D.~Metaxas.
\newblock Stackgan: Text to photo-realistic image synthesis with stacked
  generative adversarial networks.
\newblock In {\em {ICCV}}, 2017.

\bibitem{zhang2016learning}
L.~Zhang, T.~Xiang, and S.~Gong.
\newblock Learning a deep embedding model for zero-shot learning.
\newblock In {\em {CVPR}}, 2016.

\bibitem{zhang2018txt2img}
Z.~Zhang, Y.~Xie, and L.~Yang.
\newblock Photographic text-to-image synthesis with a hierarchically-nested
  adversarial network.
\newblock In {\em {CVPR}}, 2018.

\bibitem{zhang2018cardiac}
Z.~Zhang, L.~Yang, and Y.~Zheng.
\newblock Translating and segmenting multimodal medical volumes with cycle- and
  shape-consistency generative adversarial network.
\newblock In {\em {CVPR}}, 2018.

\bibitem{zhao2016energy}
J.~Zhao, M.~Mathieu, and Y.~LeCun.
\newblock Energy-based generative adversarial network.
\newblock In {\em {ICLR}}, 2017.

\end{thebibliography}
}

\end{document}